\documentclass[10pt,twocolumn,letterpaper]{article}

\usepackage{titling}
\usepackage{iccv}
\usepackage{times}
\usepackage{epsfig}
\usepackage{graphicx}
\usepackage{amsmath}
\usepackage{amssymb}

\usepackage{scalerel}
\usepackage{array}
\usepackage{multirow}
\usepackage{dsfont}
\usepackage{stfloats}
\usepackage{url}

\newcolumntype{M}{>{\centering\arraybackslash}m{.2\textwidth}}
\newcolumntype{C}[1]{>{\centering\let\newline\\\arraybackslash\hspace{0pt}}p{#1}}
\newcolumntype{R}[1]{>{\raggedleft\let\newline\\\arraybackslash\hspace{0pt}}p{#1}}
\newcolumntype{L}[1]{>{\raggedright\let\newline\\\arraybackslash\hspace{0pt}}p{#1}}

\newcommand\Tstrut{\rule{-3pt}{2.6ex}}       
\newcommand\Bstrut{\rule[-0.9ex]{-3pt}{0pt}} 
\newcommand{\TBstrut}{\rule{-3pt}{2.6ex} \rule[-0.9ex]{-2pt}{0pt}}  

\usepackage[breaklinks=true,colorlinks,bookmarks=false]{hyperref}

\iccvfinalcopy 

\def\iccvPaperID{5957} 

\ificcvfinal\fi

\begin{document}

\title{KPConv: Flexible and Deformable Convolution for Point Clouds}

\author{
Hugues Thomas$^1$
\quad
Charles R. Qi$^2$
\quad
Jean-Emmanuel Deschaud$^1$
\quad
Beatriz Marcotegui$^1$
\\
Fran\c{c}ois Goulette$^1$
\quad
Leonidas J. Guibas$^{2,3}$
\\
\\
$^1$Mines ParisTech \quad $^2$Facebook AI Research \quad $^3$Stanford University
}


\maketitle

\ificcvfinal\thispagestyle{empty}\fi

\begin{abstract}
We present Kernel Point Convolution\footnote{Project page: \textit{\urlstyle{sf}\url{https://github.com/HuguesTHOMAS/KPConv}}} (KPConv), a new design of point convolution, i.e. that operates on point clouds without any intermediate representation. The convolution weights of KPConv are located in Euclidean space by kernel points, and applied to the input points close to them. Its capacity to use any number of kernel points gives KPConv more flexibility than fixed grid convolutions. Furthermore, these locations are continuous in space and can be learned by the network. Therefore, KPConv can be extended to deformable convolutions that learn to adapt kernel points to local geometry. Thanks to a regular subsampling strategy, KPConv is also efficient and robust to varying densities. Whether they use deformable KPConv for complex tasks, or rigid KPconv for simpler tasks, our networks outperform state-of-the-art classification and segmentation approaches on several datasets. We also offer ablation studies and visualizations to provide understanding of what has been learned by KPConv and to validate the descriptive power of deformable KPConv.
\end{abstract}

\section{Introduction}

The dawn of deep learning has boosted modern computer vision with discrete convolution as its fundamental building block. This operation combines the data of local neighborhoods on a 2D grid. Thanks to this regular structure, it can be computed with high efficiency on modern hardware, but when deprived of this regular structure, the convolution operation has yet to be defined properly, with the same efficiency as on 2D grids. 

Many applications relying on such irregular data have grown with the rise of 3D scanning technologies. For example, 3D point cloud segmentation or 3D simultaneous localization and mapping rely on non-grid structured data: point clouds. A point cloud is a set of points in 3D (or higher-dimensional) space. In many applications, the points are coupled with corresponding features like colors. In this work, we will always consider a point cloud as those two elements: the points $\mathcal{P} \in \mathbb{R}^{N \times 3}$ and the features $\mathcal{F} \in \mathbb{R}^{N \times D}$. Such a point cloud is a \textit{sparse} structure that has the property to be \textit{unordered}, which makes it very different from a grid. However, it shares a common property with a grid which is essential to the definition of convolutions: it is \textit{spatially localized}. In a grid, the features are localized by their index in a matrix, while in a point cloud, they are localized by their corresponding point coordinates. Thus, the points are to be considered as structural elements, and the features as the real data.

\begin{figure}[b]
    \vspace{-3.5ex}
    \centering
    \includegraphics[width=0.98\columnwidth, keepaspectratio=true]{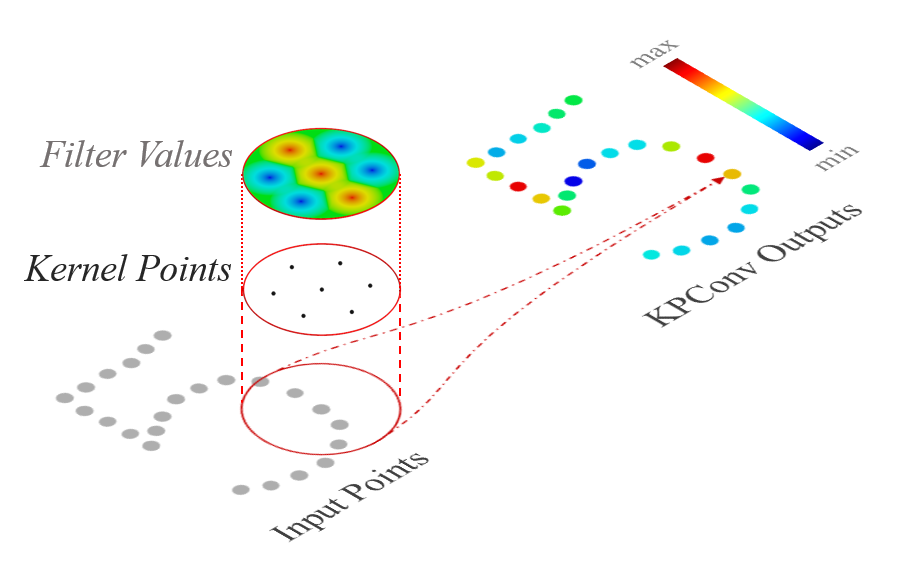}
    \caption{KPConv illustrated on 2D points. Input points with a constant scalar feature (in grey) are convolved through a KPConv that is defined by a set of kernel points (in black) with filter weights on each point.}
    \label{fig_intro}
\end{figure}

Various approaches have been proposed to handle such data, and can be grouped into different categories that we will develop in the related work section. Several methods fall into the grid-based category, whose principle is to project the sparse 3D data on a regular structure where a convolution operation can be defined more easily \cite{maturana2015voxnet, riegler2017octnet, su2018splatnet}. Other approaches use multilayer perceptrons (MLP) to process point clouds directly, following the idea proposed by \cite{zaheer2017deep, qi2017pointnet}.

More recently, some attempts have been made to design a convolution that operates directly on points \cite{atzmon2018point, xu2018spidercnn, li2018pointcnn, hua2018pointwise, hermosilla2018monte}. These methods use the \textit{spatial localization} property of a point cloud to define point convolutions with spatial kernels. They share the idea that a convolution should define a set of customizable spatial filters applied locally in the point cloud. 

This paper introduces a new point convolution operator named \textbf{Kernel Point Convolution} (KPConv). KPConv also consists of a set of local 3D filters, but overcomes previous point convolution limitations as shown in related work. KPConv is inspired by image-based convolution, but in place of kernel pixels, we use a set of kernel points to define the area where each kernel weight is applied, like shown in Figure \ref{fig_intro}. The kernel weights are thus carried by points, like the input features, and their area of influence is defined by a correlation function. The number of kernel points is not constrained, making our design very flexible. Despite the resemblance of vocabulary, our work differs from \cite{shen2018mining}, which is inspired from point cloud registration techniques, and uses kernel points without any weights to learns local geometric patterns. 

Furthermore, we propose a deformable version of our convolution \cite{dai2017deformable}, which consists of learning local shifts applied to the kernel points (see Figure \ref{fig_deform}). Our network generates different shifts at each convolution location, meaning that it can adapt the shape of its kernels for different regions of the input cloud. Our deformable convolution is not designed the same way as its image counterpart. Due to the different nature of the data, it needs a regularization to help the deformed kernels fit the point cloud geometry and avoid empty space. We use Effective Receptive Field (ERF) \cite{luo2016understanding} and ablation studies to compare rigid KPConv with deformable KPConv.

As opposed to \cite{wang2018deep, atzmon2018point, xu2018spidercnn, li2018pointcnn}, we favor radius neighborhoods instead of k-nearest-neighbors (KNN). As shown by \cite{hermosilla2018monte}, KNN is not robust in non-uniform sampling settings. The robustness of our convolution to varying densities is ensured by the combination of radius neighborhoods and regular subsampling of the input cloud \cite{thomas2018semantic}. Compared to normalization strategies \cite{hermosilla2018monte, hua2018pointwise}, our approach also alleviates the computational cost of our convolution.

In our experiments section, we show that KPConv can be used to build very deep architectures for classification and segmentation, while keeping fast training and inference times. Overall, rigid and deformable KPConv both perform very well, topping competing algorithms on several datasets. We find that rigid KPConv achieves better performances on simpler tasks, like object classification, or small segmentation datasets. Deformable KPConv thrives on more difficult tasks, like large segmentation datasets offering many object instances and greater diversity. We also show that deformable KPConv is more robust to a lower number of kernel points, which implies a greater descriptive power. Last but not least, a qualitative study of KPConv ERF shows that deformable kernels improve the network ability to adapt to the geometry of the scene objects.

\section{Related Work}

\begin{figure*}[t!]
    \vspace{-4ex}
    \centering
    \includegraphics[width=0.98\textwidth, keepaspectratio=true]{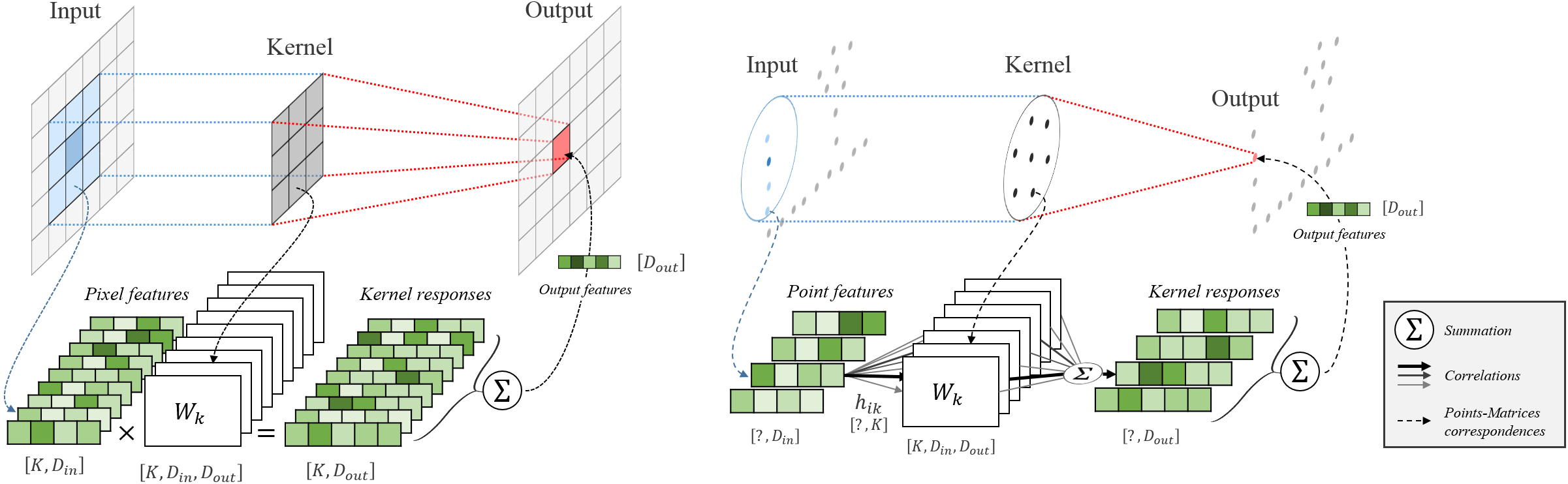}
    \caption{Comparison between an image convolution (left) and a KPConv (right) on 2D points for a simpler illustration. In the image, each pixel feature vector is multiplied by a weight matrix $(W_k)_{k<K}$ assigned by the alignment of the kernel with the image. In KPConv, input points are not aligned with kernel points, and their number can vary. Therefore, each point feature $f_i$ is multiplied by all the kernel weight matrices, with a correlation coefficient $h_{ik}$ depending on its relative position to kernel points.}
    \label{fig_KPConv}
    \vspace{-2ex}
\end{figure*}

In this section, we briefly review previous deep learning methods to analyze point clouds, paying particular attention to the methods closer to our definition of point convolutions.

\noindent
\textbf{Projection networks}. Several methods project points to an intermediate grid structure. Image-based networks are often multi-view, using a set of 2D images rendered from the point cloud at different viewpoints \cite{su2015multi, boulch2017unstructured, lawin2017deep}. For scene segmentation, these methods suffer from occluded surfaces and density variations. Instead of choosing a global projection viewpoint,  \cite{tatarchenko2018tangent} proposed projecting local neighborhoods to local tangent planes and processing them with 2D convolutions. However, this method  relies heavily on tangent estimation.

In the case of voxel-based methods, the points are projected on 3D grids in Euclidean space \cite{maturana2015voxnet, roynard2018classification, ben20183dmfv}. Using sparse structures like octrees or hash-maps allows larger grids and enhanced performances \cite{riegler2017octnet, graham20183d}, but these networks still lack flexibility as their kernels are constrained to use $3^3 = 27$ or $5^3 = 125$ voxels. Using a permutohedral lattice instead of an Euclidean grid reduces the kernel to $15$ lattices \cite{su2018splatnet}, but this number is still constrained, while KPConv allows any number of kernel points. Moreover, avoiding intermediate structures should make the design of more complex architectures like instance mask detector or generative models more straightforward in future works.

\noindent
\textbf{Graph convolution networks}. The definition of a convolution operator on a graph has been addressed in different ways. A convolution on a graph can be computed as a multiplication on its spectral representation \cite{defferrard2016convolutional, yi2017syncspeccnn}, or it can focus on the surface represented by the graph \cite{masci2015geodesic, bronstein2017geometric, simonovsky2017dynamic, monti2017geometric}. Despite the similarity between point convolutions and the most recent graph convolutions \cite{verma2018feastnet, wang2018dynamic}, the latter learn filters on edge relationships instead of points relative positions. In other words, a graph convolution combines features on local surface patches, while being invariant to the deformations of those patches in Euclidean space. In contrast, KPConv combines features locally according to the 3D geometry, thus capturing the deformations of the surfaces.

\noindent
\textbf{Pointwise MLP networks}. PointNet \cite{qi2017pointnet} is considered a milestone in point cloud deep learning. This network uses a shared MLP on every point individually followed by a global max-pooling. The shared MLP acts as a set of learned spatial encodings and the global signature of the input point cloud is computed as the maximal response among all the points for each of these encodings. The network's performances are limited because it does not consider local spatial relationships in the data. Following PointNet, some hierarchical architectures have been developed to aggregate local neighborhood information with MLPs \cite{qi2017pointnet++, li2018so, liu2018point2sequence}.

As shown by \cite{wang2018deep, li2018pointcnn, hermosilla2018monte}, the kernel of a point convolution can be implemented with a MLP, because of its ability to approximate any continuous function. However, using such a representation makes the convolution operator more complex and the convergence of the network harder. In our case, we define an explicit convolution kernel, like image convolutions, whose weights are directly learned, without the intermediate representation of a MLP. Our design also offers a straightforward deformable version, as offsets can directly be applied to kernel points.

\noindent
\textbf{Point convolution networks}. Some very recent works also defined explicit convolution kernels for points, but KPConv stands out with unique design choices.

Pointwise CNN \cite{hua2018pointwise} locates the kernel weights with voxel bins, and thus lacks flexibility like grid networks. Furthermore, their normalization strategy burdens their network with unnecessary computations, while KPConv subsampling strategy alleviates both varying densities and computational cost.

SpiderCNN \cite{xu2018spidercnn} defines its kernel as a family of polynomial functions applied with a different weight for each neighbor. The weight applied to a neighbor depends on the neighbor's distance-wise order, making the filters spatially inconsistent. By contrast, KPConv weights are located in space and its result is invariant to point order.

Flex-convolution \cite{groh2018flex} uses linear functions to model its kernel, which could limit its representative power. It also uses KNN, which is not robust to varying densities as discussed above.

PCNN \cite{atzmon2018point} design is the closest to KPConv. Its definition also uses points to carry kernel weights, and a correlation function. However, this design is not scalable because it does not use any form of neighborhood, making the convolution computations quadratic on the number of points. In addition, it uses a Gaussian correlation where KPConv uses a simpler linear correlation, which helps gradient backpropagation when learning deformations \cite{dai2017deformable}.

We show that KPConv networks outperform all comparable networks in the experiments section. Furthermore, to the best of our knowledge, none of the previous works experimented a spatially deformable point convolution.

\section{Kernel Point Convolution}

\subsection{A Kernel Function Defined by Points}

Like previous works, KPConv can be formulated with the general definition of a point convolution (Eq. \ref{eq:1}), inspired by image convolutions. For the sake of clarity, we call $x_i$ and $f_i$ the points from $\mathcal{P} \in \mathbb{R}^{N \times 3}$ and their corresponding features from $\mathcal{F} \in \mathbb{R}^{N \times D}$. The general point convolution of $\mathcal{F}$ by a kernel $g$ at a point $x \in \mathbb{R}^3$ is defined as:

\begin{equation} \label{eq:1}
    (\mathcal{F}*g)(x) = \sum_{x_i \in \mathcal{N}_x} g(x_i-x)f_i
\end{equation}

We stand with \cite{hermosilla2018monte} advising radius neighborhoods to ensure robustness to varying densities, therefore, $\mathcal{N}_x = 
    \left\lbrace
        \begin{array}{c|c}  
            x_i \in \mathcal{P} &  \left\Vert x_i - x \right\Vert \leqslant r 
        \end{array} 
    \right\rbrace$ with $r \in \mathbb{R}$ being the chosen radius. In addition, \cite{thomas2018semantic} showed that hand-crafted 3D point features offer a better representation when computed with radius neighborhoods than with KNN. We believe that having a consistent spherical domain for the function $g$ helps the network to learn meaningful representations. 
    
The crucial part in Eq. \ref{eq:1} is the definition of the kernel function $g$, which is where KPConv singularity lies. $g$ takes the neighbors positions centered on $x$ as input. We call them $y_i = x_i - x$ in the following. As our neighborhoods are defined by a radius $r$, the domain of definition of $g$ is the ball $\mathcal{B}_r^3 = \left\lbrace y \in \mathbb{R}^3  \: | \: \left\Vert y \right\Vert \leqslant r \right\rbrace$. Like image convolution kernels (see Figure \ref{fig_KPConv} for a detailed comparison between image convolution and KPConv), we want $g$ to apply different weights to different areas inside this domain. There are many ways to define areas in 3D space, and points are the most intuitive as features are also localized by them. Let $\left\lbrace \widetilde{x}_k \: | \: k < K \right\rbrace \subset \mathcal{B}_r^3$ be the kernel points and $\left\lbrace W_k \: | \: k < K \right\rbrace \subset \mathbb{R}^{D_{in} \times D_{out}}$ be the associated weight matrices that map features from dimension $D_{in}$ to $D_{out}$. We define the kernel function $g$ for any point $y_i\in\mathcal{B}_r^3$ as :

\begin{equation} \label{eq:2}
    g(y_i) = \sum_{k<K} h\left(y_i,  \widetilde{x}_k\right) W_k
\end{equation}

\noindent
where $h$ is the correlation between $\widetilde{x}_k$ and $y_i$, that should be higher when $\widetilde{x}_k$ is closer to $y_i$. Inspired by the bilinear interpolation in \cite{dai2017deformable}, we use the linear correlation:

\begin{equation} \label{eq:3}
    h\left(y_i,  \widetilde{x}_k\right) = \max\left(0, 1 - \frac{\left\Vert y_i - \widetilde{x}_k \right\Vert}{\sigma}\right)
\end{equation}

\noindent
where $\sigma$ is the influence distance of the kernel points, and will be chosen according to the input density (see Section \ref{section_33}). Compared to a gaussian correlation, which is used by \cite{atzmon2018point}, linear correlation is a simpler representation. We advocate this simpler correlation to ease gradient backpropagation when learning kernel deformations. A parallel can be drawn with rectified linear unit, which is the most popular activation function for deep neural networks, thanks to its efficiency for gradient backpropagation.

\subsection{Rigid or Deformable Kernel}

Kernel point positions are critical to the convolution operator. Our rigid kernels in particular need to be arranged regularly to be efficient. As we claimed that one of the KPConv strengths is its flexibility, we need to find a regular disposition for any $K$. We chose to place the kernel points by solving an optimization problem where each point applies a repulsive force on the others. The points are constrained to stay in the sphere with an attractive force, and one of them is constrained to be at the center. We detail this process and show some regular dispositions in the supplementary material. Eventually, the surrounding points are rescaled to an average radius of $1.5\sigma$, ensuring a small overlap between each kernel point area of influence and a good space coverage.

With properly initialized kernels, the rigid version of KPConv is extremely efficient, in particular when given a large enough $K$ to cover the spherical domain of $g$. However it is possible to increase its capacity by learning the kernel point positions. The kernel function $g$ is indeed differentiable with respect to $\widetilde{x}_k$, which means they are learnable parameters. We could consider learning one global set of $\left\lbrace\widetilde{x}_k\right\rbrace$ for each convolution layer, but it would not bring more descriptive power than a fixed regular disposition. Instead the network generates a set of $K$ shifts $\Delta(x)$ for every convolution location $x \in \mathbb{R}^3$ like \cite{dai2017deformable} and define deformable KPConv as:

\begin{equation} \label{eq:4}
    (\mathcal{F}*g)(x) = \sum_{x_i \in \mathcal{N}_x} g_\mathit{deform}(x-x_i, \Delta(x))f_i
\end{equation}
\begin{equation} \label{eq:5}
    g_\mathit{deform}(y_i, \Delta(x)) = \sum_{k<K} h\left(y_i,  \widetilde{x}_k + \Delta_k(x) \right) W_k
\end{equation}

\begin{figure}[t!]
    \centering
    \includegraphics[width=0.98\columnwidth, keepaspectratio=true]{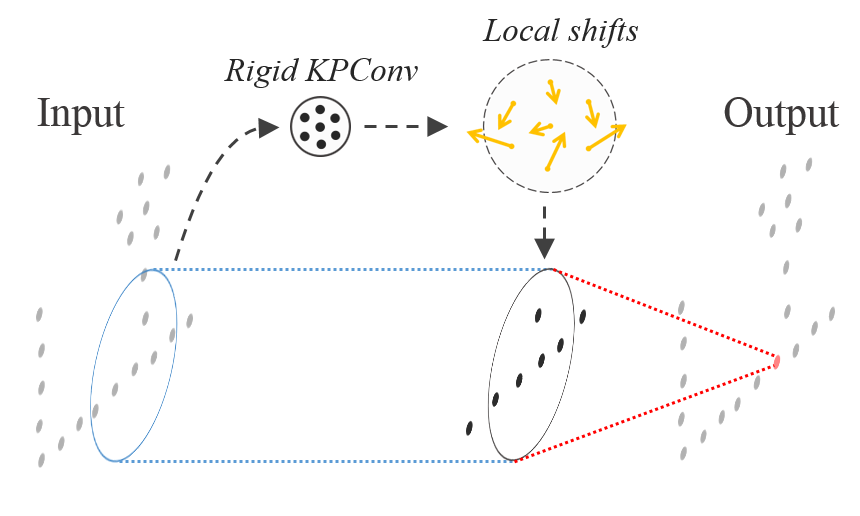}
    \caption{Deformable KPConv illustrated on 2D points.}
    \label{fig_deform}
    \vspace{-2ex}
\end{figure}

We define the offsets $\Delta_k(x)$ as the output of a rigid KPConv mapping $D_{in}$  input features to $3K$ values, as shown in Figure \ref{fig_deform}. During training, the network learns the rigid kernel generating the shifts and the deformable kernel generating the output features simultaneously, but the learning rate of the first one is set to $0.1$ times the global network learning rate.

Unfortunately, this straightforward adaptation of image deformable convolutions does not fit point clouds. In practice, the kernel points end up being pulled away from the input points. These kernel points are lost by the network, because the gradients of their shifts $\Delta_k(x)$ are null when no neighbors are in their influence range. More details on these ``lost" kernel points are given in the supplementary. To tackle this behaviour, we propose a ``fitting'' regularization loss which penalizes the distance between a kernel point and its closest neighbor among the input neighbors. In addition, we also add a ``repulsive'' regularization loss between all pair off kernel points when their influence area overlap, so that they do not collapse together. As a whole our regularization loss for all convolution locations $x \in \mathbb{R}^3$ is:

\begin{equation} \label{eq:6}
    \mathcal{L}_\mathit{reg} = \sum_{x} \mathcal{L}_\mathit{fit}(x) + \mathcal{L}_\mathit{rep}(x)
    \vspace{-2ex}
\end{equation}

\begin{equation} \label{eq:7}
    \mathcal{L}_\mathit{fit}(x) = \sum_{k < K} \min_{y_i}  \left(\frac{\left\Vert y_i - (\widetilde{x}_k + \Delta_k(x)) \right\Vert}{\sigma}\right)^2
    \vspace{-2ex}
\end{equation}

\begin{equation} \label{eq:8}
    \mathcal{L}_\mathit{rep}(x) = \sum_{k < K}\sum_{l\neq k} h\left(\widetilde{x}_k + \Delta_k(x), \widetilde{x}_l + \Delta_l(x)\right)^2
    \vspace{-1ex}
\end{equation}

With this loss, the network generates shifts that fit the local geometry of the input point cloud. We show this effect in the supplementary material.

\subsection{Kernel Point Network Layers}
\label{section_33}

This section elucidates how we effectively put the KPConv theory into practice. For further details, we have released our code using Tensorflow library.

\noindent
\textbf{Subsampling to deal with varying densities}. As explained in the introduction, we use a subsampling strategy to control the density of input points at each layer. To ensure a spatial consistency of the point sampling locations, we favor grid subsampling. Thus, the support points of each layer, carrying the features locations, are chosen as barycenters of the original input points contained in all non-empty grid cells. 

\noindent
\textbf{Pooling layer}. To create architectures with multiple layer scales, we need to reduce the number of points progressively. As we already have a grid subsampling, we double the cell size at every pooling layer, along with the other related parameters, incrementally increasing the receptive field of KPConv. The features pooled at each new location can either be obtained by a max-pooling or a KPConv. We use the latter in our architectures and call it ``strided KPConv", by analogy to the image strided convolution.

\noindent
\textbf{KPConv layer}. Our convolution layer takes as input the points $\mathcal{P} \in \mathbb{R}^{N \times 3}$, their corresponding features $\mathcal{F} \in \mathbb{R}^{N \times D_{in}}$, and the matrix of neighborhood indices $\mathfrak{N} \in {[\![1,N]\!]}^{N' \times n_{max}}$. $N'$ is the number of locations where the neighborhoods are computed, which can be different from $N$ (in the case of ``strided" KPConv). The neighborhood matrix is forced to have the size of the biggest neighborhood $n_{max}$. Because most of the neighborhoods comprise less than $n_{max}$ neighbors, the matrix $\mathfrak{N}$ thus contains unused elements. We call them “shadow” neighbors, and they are ignored during the convolution computations.

\noindent
\textbf{Network parameters}. Each layer $j$ has a cell size ${dl}_j$ from which we infer other parameters. The kernel points influence distance is set as equal to $\sigma_j = \Sigma \times {dl}_j$. For rigid KPConv, the convolution radius is automatically set to $2.5 \, \sigma_j$ given that the average kernel point radius is $1.5 \, \sigma_j$. For deformable KPConv, the convolution radius can be chosen as $r_j = \rho \times {dl}_j$. $\Sigma$ and $\rho$ are proportional coefficients set for the whole network. Unless stated otherwise, we will use the following set of parameters, chosen by cross validation, for all experiments: $K=15$, $\Sigma = 1.0$ and $\rho = 5.0$. The first subsampling cell size ${dl}_0$ will depend on the dataset and, as stated above, ${dl}_{j+1} = 2 * {dl}_j$.

\subsection{Kernel Point Network Architectures}

Combining analogy with successful image networks and empirical studies, we designed two network architectures for the classification and the segmentation tasks. Diagrams detailing both architectures are available in the supplementary material.

\noindent
\textbf{KP-CNN} is a 5-layer classification convolutional network. Each layer contains two convolutional blocks, the first one being strided except for the first layer. Our convolutional blocks are designed like bottleneck ResNet blocks \cite{he2016deep} with a KPConv replacing the image convolution, batch normalization and leaky ReLu activation. After the last layer, the features are aggregated by a global average pooling and processed by the fully connected and softmax layers like in an image CNN. For the results with deformable KPConv, we only use deformable kernels in the last 5 KPConv blocks (see architecture details in the supplementary material).

\noindent
\textbf{KP-FCNN} is a fully convolutional network for segmentation. The encoder part is the same as in KP-CNN, and the decoder part uses nearest upsampling to get the final pointwise features. Skip links are used to pass the features between intermediate layers of the encoder and the decoder. Those features are concatenated to the upsampled ones and processed by a unary convolution, which is the equivalent of a $1 \times 1$ convolution in image or a shared MLP in PointNet. It is possible to replace the nearest upsampling operation by a KPConv, in the same way as the strided KPConv, but it does not lead to a significant improvement of the performances.

\section{Experiments}

\subsection{3D Shape Classification and Segmentation}

\noindent
\textbf{Data}. First, we evaluate our networks on two common model datasets. We use ModelNet40 \cite{wu20153d} for classification and ShapenetPart \cite{yi2016scalable} for part segmentation. ModelNet40 contains 12,311 meshed CAD models from 40 categories. ShapenetPart is a collection of 16,681 point clouds from 16 categories, each with 2-6 part labels. For benchmarking purpose, we use data provided by \cite{qi2017pointnet++}. In both cases, we follow standard train/test splits and rescale objects to fit them into a unit sphere (and consider units to be meters for the rest of this experiment). We ignore normals because they are only available for artificial data.

\noindent
\textbf{Classification task}. We set the first subsampling grid size to ${dl}_0=2\mathrm{cm}$. We do not add any feature as input; each input point is assigned a constant feature equal to 1, as opposed to empty space which can be considered as 0. This constant feature encodes the geometry of the input points. Like \cite{atzmon2018point}, our augmentation procedure consists of scaling, flipping and perturbing the points. In this setup, we are able to process $2.9$ batches of 16 clouds per second on an Nvidia Titan Xp. Because of our subsampling strategy, the input point clouds do not all have the same number of points, which is not a problem as our networks accept variable input point cloud size. On average, a ModelNet40 object point cloud comprises 6,800 points in our framework. The other training parameters are detailed in the supplementary material, along with the architecture details. We also include the number of parameters and the training/inference speeds for both rigid and deformable KPConv.

As shown on Table \ref{Table_ModelNet40}, our networks outperform other state-of-the-art methods using only points (we do not take into account methods using normals as additional input). We also notice that rigid KPConv performances are slightly better. We suspect that it can be explained by the task simplicity. If deformable kernels add more descriptive power, they also increase the overall network complexity, which can disturb the convergence or lead to overfitting on simpler tasks like this shape classification. 

\noindent
\textbf{Segmentation task}. For this task, we use KP-FCNN architecture with the same parameters as in the classification task, adding the positions $(x, y, z)$ as additional features to the constant 1, and using the same augmentation procedure. We train a single network with multiple heads to segment the parts of each object class. The clouds are smaller (2,300 points on average), and we can process $4.1$ batches of 16 shapes per second. Table \ref{Table_ModelNet40} shows the instance average, and the class average mIoU. We detail each class mIoU in the supplementary material. KP-FCNN outperforms all other algorithms, including those using additional inputs like images or normals. Shape segmentation is a more difficult task than shape classification, and we see that KPConv has better performances with deformable kernels.

\begin{table}[!t]
\setlength\tabcolsep{0.5pt}
\begin{center}
\begin{tabular}{ L{3.0cm} C{2.0cm} C{1.5cm} C{1.5cm}}
\hline
 & ModelNet40 & \multicolumn{2}{c}{ShapeNetPart} \TBstrut\\
\hline
Methods & OA & mcIoU & mIoU \TBstrut\\
\hline
SPLATNet \cite{su2018splatnet}	    & -         & $83.7$	& $85.4$    \Tstrut\\
SGPN \cite{wang2018sgpn}            & -         & $82.8$	& $85.8$	\\
3DmFV-Net \cite{graham20183d}	    & $91.6$    & $81.0$	& $84.3$	\\
SynSpecCNN \cite{yi2017syncspeccnn}	& -         & $82.0$	& $84.7$	\\
RSNet \cite{huang2018recurrent}     & -         & $81.4$    & $84.9$    \\
SpecGCN \cite{wang2018local}        & $91.5$    & -         & $85.4$    \\
PointNet++ \cite{qi2017pointnet++}	& $90.7$    & $81.9$	& $85.1$	\\
SO-Net  \cite{li2018so}	            & $90.9$    & $81.0$	& $84.9$    \\
PCNN by Ext \cite{atzmon2018point}	& $92.3$    & $81.8$	& $85.1$	\\
SpiderCNN \cite{xu2018spidercnn}    & $90.5$    & $82.4$	& $85.3$	\\
MCConv \cite{hermosilla2018monte}   & $90.9$    & -         & $85.9$    \\
FlexConv \cite{groh2018flex}        & $90.2$    & $84.7$    & $85.0$    \\
PointCNN \cite{li2018pointcnn}	    & $92.2$    & $84.6$	& $86.1$	\\
DGCNN \cite{wang2018dynamic}        & $92.2$    & $85.0$	& $84.7$    \\
SubSparseCNN \cite{graham20183d}	& -         & $83.3$	& $86.0$	\Bstrut\\
\hline
KPConv \textit{rigid} & $\mathbf{92.9}$ & $85.0$	& $86.2$ \Tstrut\\
KPConv \textit{deform} & $92.7$ & $\mathbf{85.1}$ & $\mathbf{86.4}$ \Bstrut\\
\hline
\end{tabular}
\end{center}
\caption{3D Shape Classification and Segmentation results. For generalizability to real data, we only consider scores obtained without shape normals on ModelNet40 dataset. The metrics are overall accuracy (OA) for Modelnet40, class average IoU (mcIoU) and instance average IoU (mIoU) for ShapeNetPart.}
\label{Table_ModelNet40}
\vspace{-3ex}
\end{table}

\subsection{3D Scene Segmentation}
\label{section_42}

\noindent
\textbf{Data}. Our second experiment shows how our segmentation architecture generalizes to real indoor and outdoor data. To this end, we chose to test our network on 4 datasets of different natures. Scannet \cite{dai2017scannet}, for indoor cluttered scenes, S3DIS \cite{armeni20163d}, for indoor large spaces, Semantic3D \cite{hackel2017semantic3d}, for outdoor fixed scans, and Paris-Lille-3D \cite{roynard2018paris}, for outdoor mobile scans. Scannet contains 1,513 small training scenes and 100 test scenes for online benchmarking, all annotated with 20 semantic classes. S3DIS covers six large-scale indoor areas from three different buildings for a total of 273 million points annotated with 13 classes. Like \cite{tchapmi2017segcloud}, we advocate the use of Area-5 as test scene to better measure the generalization ability of our method. Semantic3D is an online benchmark comprising several fixed lidar scans of different outdoor scenes. More than 4 billion points are annotated with 8 classes in this dataset, but they mostly cover ground, building or vegetation and there are fewer object instances than in the other datasets. We favor the \textit{reduced-8} challenge because it is less biased by the objects close to the scanner. Paris-Lille-3D contains more than 2km of streets in 4 different cities and is also an online benchmark. The 160 million points of this dataset are annotated with 10 semantic classes.

\noindent
\textbf{Pipeline for real scene segmentation}. The 3D scenes in these datasets are too big to be segmented as a whole. Our KP-FCNN architecture is used to segment small subclouds contained in spheres. At training, the spheres are picked randomly in the scenes. At testing, we pick spheres regularly in the point clouds but ensure each point is tested multiple times by different sphere locations. As in a voting scheme on model datasets, the predicted probabilities for each point are averaged. When datasets are colorized, we use the three color channels as features. We still keep the constant 1 feature to ensure black/dark points are not ignored. To our convolution, a point with all features equal to zero is equivalent to empty space. The input sphere radius is chosen as $50 \times {dl}_0$ (in accordance to Modelnet40 experiment).

\begin{table}[!t]
\setlength\tabcolsep{0.5pt}
\begin{center}
\begin{tabular}{ L{2.8cm} C{1.3cm}  C{1.3cm} C{1.3cm} C{1.3cm}}
\hline
Methods & Scannet & Sem3D & S3DIS & PL3D \TBstrut\\
\hline

Pointnet \cite{qi2017pointnet}	            & - & - & $41.1$ & - \Tstrut\\
Pointnet++ \cite{qi2017pointnet++}	        & $33.9$ & - & - & - \\
SnapNet \cite{boulch2017unstructured}	    & - & $59.1$ & - & - \\
SPLATNet \cite{su2018splatnet}	            & $39.3$ & - & - & - \\
SegCloud \cite{tchapmi2017segcloud}	        & - & $61.3$ & $48.9$ & -\\
RF\_MSSF \cite{thomas2018semantic}	        & - & $62.7$ & $49.8$ & $56.3$ \\
Eff3DConv \cite{zhang2018efficient}	        & - & - & $51.8$ & -\\
TangentConv \cite{tatarchenko2018tangent}	& $43.8$ & - & $52.6$ & -\\
MSDVN \cite{roynard2018classification}      & - & $65.3$ & $54.7$ & $66.9$	\\
RSNet \cite{huang2018recurrent}             & - & - & $56.5$ & -	\\
FCPN \cite{rethage2018fully}	            & $44.7$ & - & - & - \\
PointCNN \cite{li2018pointcnn}	            & $45.8$ & - & $57.3$ & - \\
PCNN \cite{atzmon2018point}	                & $49.8$ & - & - & -\\
SPGraph \cite{landrieu2018large}	        & - & $73.2$ & $58.0$ & -	\\
ParamConv \cite{wang2018deep}	            & - & - & $58.3$ & -  \\
SubSparseCNN \cite{graham20183d}	        & $\mathbf{72.5}$ & - & - & - \Bstrut\\
\hline
KPConv \textit{rigid} & $68.6$ & $\mathbf{74.6}$	& $65.4$ & $72.3$ \Tstrut\\
KPConv \textit{deform} & $68.4$ & $73.1$	& $\mathbf{67.1}$ & $\mathbf{75.9}$ \Bstrut\\
\hline
\end{tabular}
\end{center}
\caption{3D scene segmentation scores (mIoU). Scannet, Semantic3D and Paris-Lille-3D (PL3D) scores are taken from their respective online benchmarks (reduced-8 challenge for Semantic3D). S3DIS scores are given for Area-5 (see supplementary material for k-fold).}
\label{Table_Semantic3D}
\vspace{-3ex}
\end{table}

\noindent
\textbf{Results}. Because outdoor objects are larger than indoor objects, we use ${dl}_0=6 \mathrm{cm}$ on Semantic3D and Paris-Lille-3D, and ${dl}_0=4 \mathrm{cm}$ on Scannet and S3DIS. As shown in Table \ref{Table_Semantic3D}, our architecture ranks second on Scannet and outperforms all other segmentation architectures on the other datasets. Compared to other point convolution architectures \cite{atzmon2018point, li2018pointcnn, wang2018deep}, KPConv performances exceed previous scores by 19 mIoU points on Scannet and 9 mIoU points on S3DIS. SubSparseCNN score on Scannet was not reported in their original paper \cite{graham20183d}, so it is hard to compare without knowing their experimental setup. We can notice that, in the same experimental setup on ShapeNetPart segmentation, KPConv outperformed SubSparseCNN by nearly 2 mIoU points. 

Among these 4 datasets, KPConv deformable kernels improved the results on Paris-Lille-3D and S3DIS while the rigid version was better on Scannet and Semantic3D. If we follow our assumption, we can explain the lower scores on Semantic3D by the lack of diversity in this dataset. Indeed, despite comprising 15 scenes and 4 billion points, it contains a majority of ground, building and vegetation points and a few real objects like car or pedestrians. Although this is not the case of Scannet, which comprises more than 1,500 scenes with various objects and shapes, our validation studies are not reflected by the test scores on this benchmark. We found that the deformable KPConv outperformed its rigid counterpart on several different validation sets (see Section \ref{section_43}). As a conclusion, these results show that the descriptive power of deformable KPConv is useful to the network on large and diverse datasets. We believe KPConv could thrive on larger datasets because its kernel combines a strong descriptive power (compared to other simpler representations, like the linear kernels of \cite{groh2018flex}), and great learnability (the weights of MLP convolutions like \cite{li2018pointcnn, wang2018deep} are more complex to learn). An illustration of segmented scenes on Semantic3D and S3DIS is shown in Figure \ref{fig_results}. More results visualizations are provided in the supplementary material. 

\subsection{Ablation Study}
\label{section_43}

\begin{figure}[b]
    \vspace{-3ex}
    \centering
    \includegraphics[width=0.98\columnwidth, keepaspectratio=true]{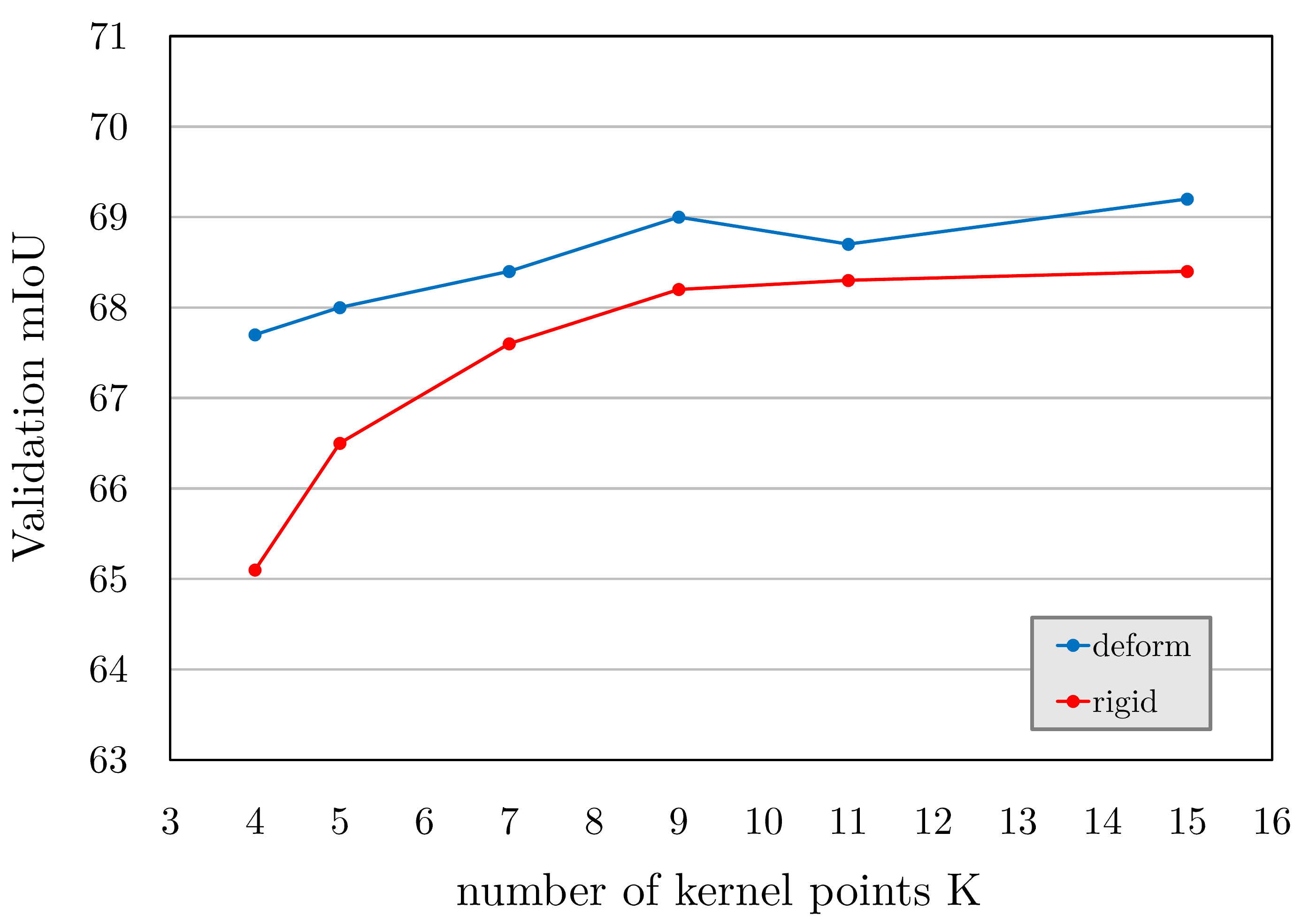}
    \caption{Ablation study on Scannet validation set. Evolution of the mIoU when reducing the number of kernel points.}
    \label{fig_ablation}
\end{figure}

We conduct an ablation study to support our claim that deformable KPConv has a stronger descriptive power than rigid KPConv. The idea is to impede the capabilities of the network, in order to reveal the real potential of deformable kernels. We use Scannet dataset (same parameters as before) and use the official validation set, because the test set cannot be used for such evaluations. As depicted in Figure \ref{fig_ablation}, the deformable KPConv only loses $1.5\%$ mIoU when restricted to 4 kernel points. In the same configuration, the rigid KPConv loses $3.5\%$ mIoU.

As stated in Section \ref{section_42}, we can also see that deformable KPConv performs better than rigid KPConv with 15 kernel points. Although it is not the case on the test set, we tried different validation sets that confirmed the superior performances of deformable KPConv. This is not surprising as we obtained the same results on S3DIS. Deformable KPConv seem to thrive on indoor datasets, which offer more diversity than outdoor datasets. To understand why, we need to go beyond numbers and see what is effectively learned by the two versions of KPConv. 

\begin{figure}[t]
    \centering
    \includegraphics[width=0.98\columnwidth, keepaspectratio=true]{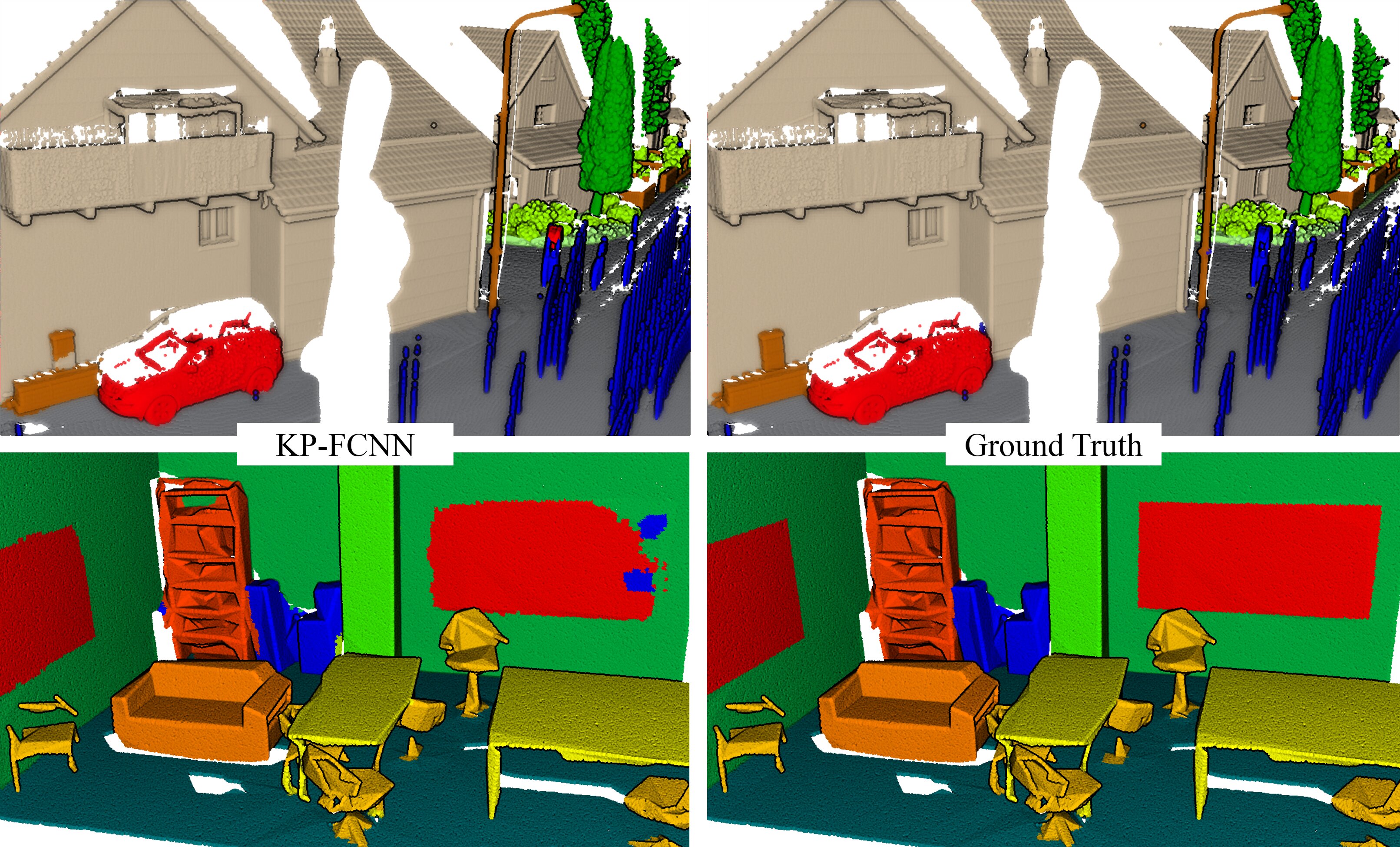}
    \caption{Outdoor and Indoor scenes, respectively from Semantic3D and S3DIS, classified by KP-FCNN with deformable kernels.}
    \label{fig_results}
    \vspace{-3ex}
\end{figure}

\begin{figure}[t!]
    \centering
    \includegraphics[width=0.98\columnwidth, keepaspectratio=true]{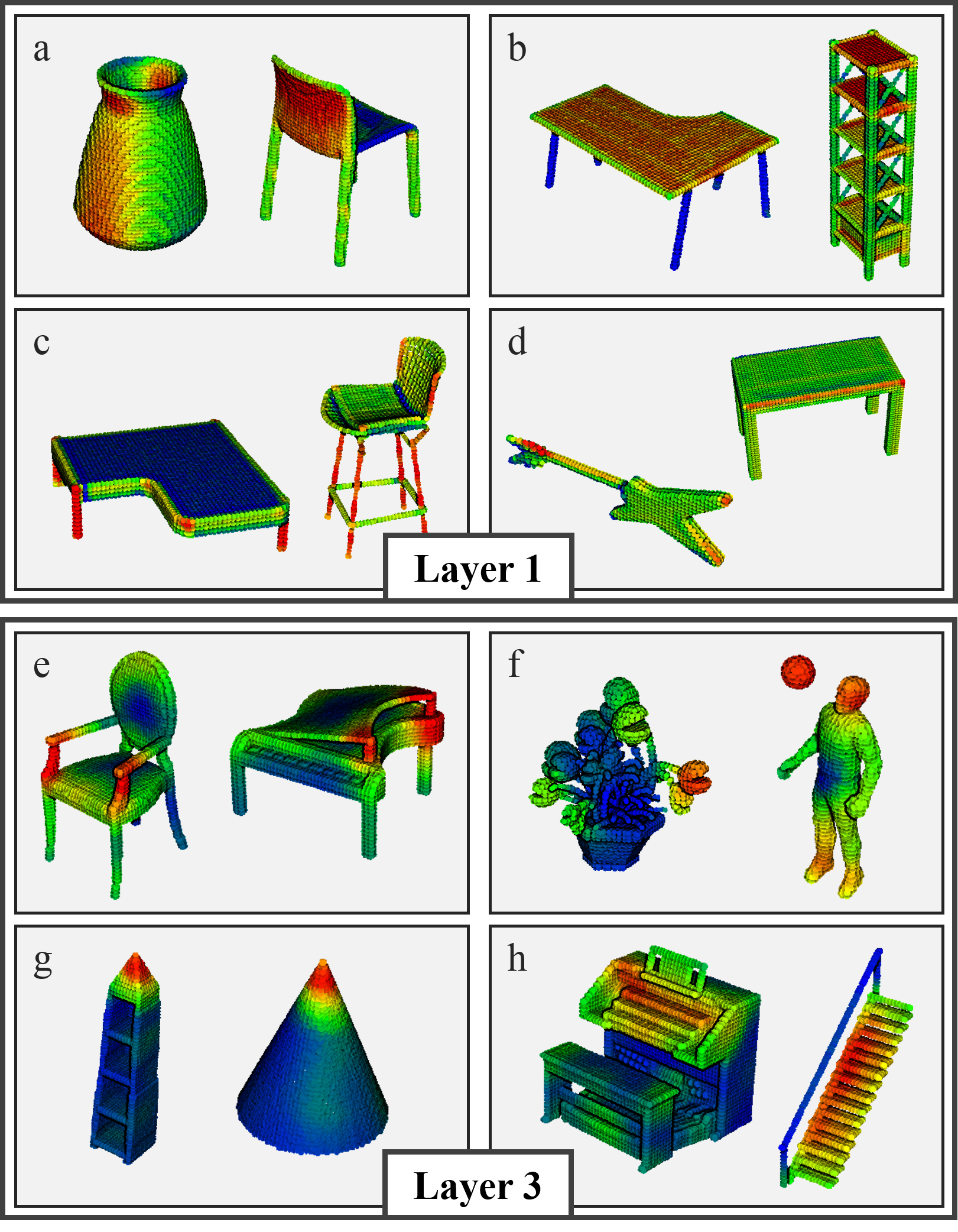}
    \caption{Low and high level features learned in KP-CNN. Each feature is displayed on 2 input point clouds taken from ModelNet40. High activations are in red and low activations in blue.}
    \label{fig_features}
    \vspace{-3ex}
\end{figure}

\subsection{Learned Features and Effective Receptive Field}

To achieve a deeper understanding of KPConv, we offer two insights of the learning mechanisms. 

\noindent
\textbf{Learned features.} Our first idea was to visualize the features learned by our network. In this experiment, we trained KP-CNN on ModelNet40 with rigid KPConv. We added random rotation augmentations around vertical axis to increase the input shape diversity. Then we visualize each learned feature by coloring the points according to their level of activation for this features. In Figure \ref{fig_features}, we chose input point clouds maximizing the activation for different features at the first and third layer. For a cleaner display, we projected the activations from the layer subsampled points to the original input points. We observe that, in its first layer, the network is able to learn low-level features like vertical/horizontal planes (a/b), linear structures (c), or corners (d). In the later layers, the network detects more complex shapes like small buttresses (e), balls (f), cones (g), or stairs (h). However, it is difficult to see a difference between rigid and deformable KPConv. This tool is very useful to understand what KPConv can learn in general, but we need another one to compare the two versions.

\begin{figure}[t!]
    \centering
    \includegraphics[width=0.98\columnwidth, keepaspectratio=true]{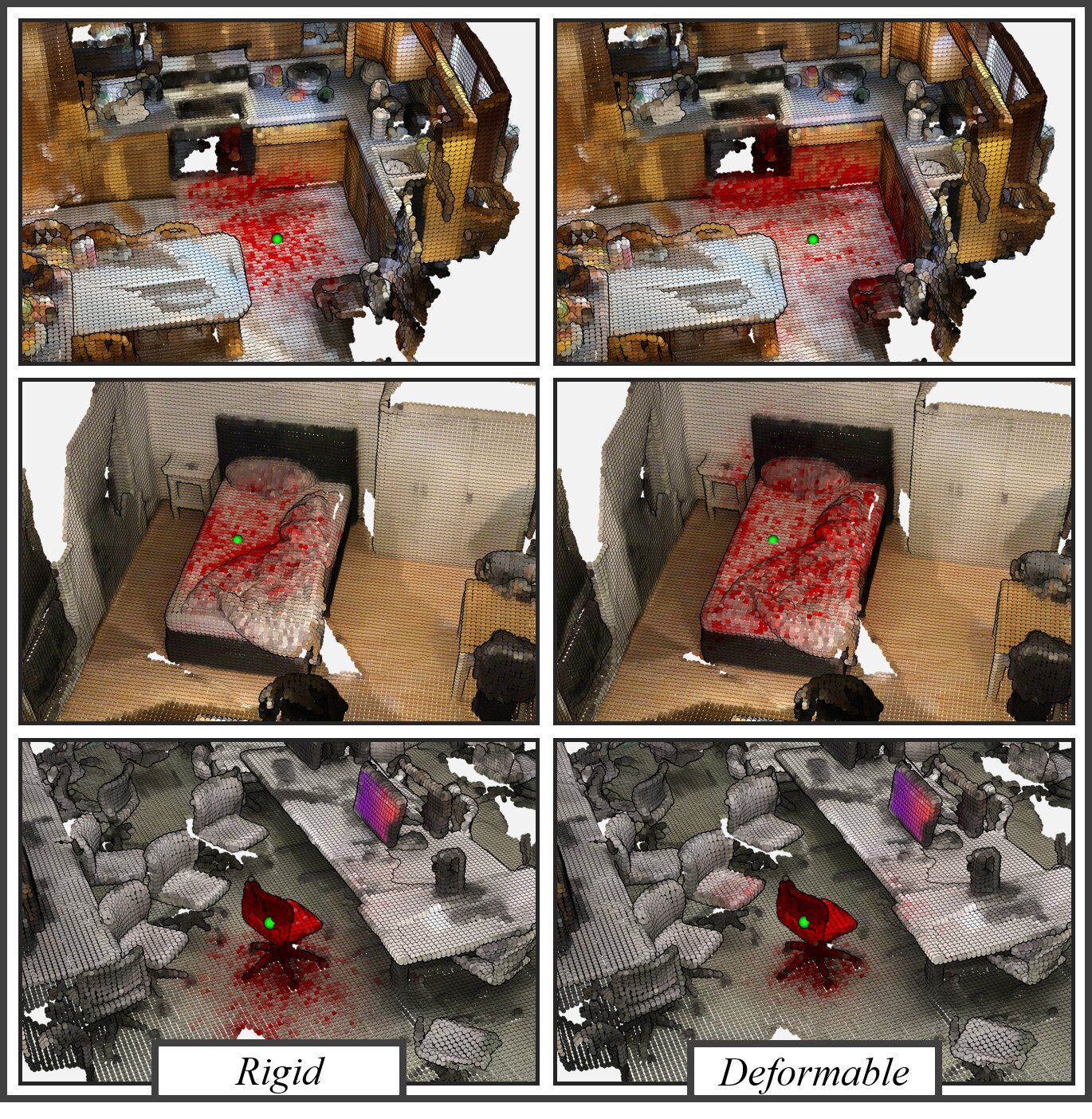}
    \caption{KPConv ERF at layer 4 of KP-FCNN, trained on Scannet. The green dots represent the ERF centers. ERF values are merged with scene colors as red intensity. The more red a point is, the more influence it has on the green point features.}
    \label{fig_ERF}
    \vspace{-3ex}
\end{figure}

\noindent
\textbf{Effective Receptive Field.} To apprehend the differences between the representations learned by rigid and deformable KPConv, we can compute its Effective Receptive Field (ERF) \cite{luo2016understanding} at different locations. The ERF is a measure of the influence that each input point has on the result of a KPConv layer at a particular location. It is computed as the gradient of KPConv responses at this particular location with respect to the input point features. As we can see in Figure \ref{fig_ERF}, the ERF varies depending on the object it is centered on. We see that rigid KPConv ERF has a relatively consistent range on every type of object, whereas deformable KPConv ERF seems to adapt to the object size. Indeed, it covers the whole bed, and concentrates more on the chair that on the surrounding ground. When centered on a flat surface, it also seems to ignore most of it and reach for further details in the scene. This adaptive behavior shows that deformable KPConv improves the network ability to adapt to the geometry of the scene objects, and explains the better performances on indoor datasets.

\section{Conclusion}

In this work, we propose KPConv, a convolution that operates on point clouds. KPConv takes radius neighborhoods as input and processes them with weights spatially located by a small set of kernel points. We define a deformable version of this convolution operator that learns local shifts effectively deforming the convolution kernels to make them fit the point cloud geometry. Depending on the diversity of the datasets, or the chosen network configuration, deformable and rigid KPConv are both valuable, and our networks brought new state-of-the-art performances for nearly every tested dataset. We release our source code, hoping to help further research on point cloud convolutional architectures. Beyond the proposed classification and segmentation networks, KPConv can be used in any other application addressed by CNNs. We believe that deformable convolutions can thrive in larger datasets or challenging tasks such as object detection, lidar flow computation, or point cloud completion.

\noindent
\textbf{Acknowledgement.} The authors gratefully acknowledge
the support of ONR MURI grant N00014-13-1-0341, NSF grant IIS-1763268, a Vannevar Bush Faculty Fellowship, and a gift from the Adobe and Autodesk corporations.

\clearpage

{\small
\bibliographystyle{ieee_fullname}
\bibliography{egbib}
}


\clearpage


\pretitle{\vspace{10ex}}
\posttitle{\vspace{10ex}}
\title{\textit{Supplementary Material for} \\ KPConv: Flexible and Deformable Convolution for Point Clouds}
\preauthor{}
\postauthor{}
\author{}

\maketitle

\renewcommand{\thesection}{\Alph{section}}
\setcounter{section}{0}


\begin{abstract}
\noindent
This supplementary document is organized as follows:
\begin{itemize}
    \itemsep 0ex 
    \item Sec.\,\ref{sec:A} details our network architectures, the training parameters, and compares the model sizes and speeds.
    \item Sec.\,\ref{sec:B} presents the kernel point initialization method.
    \item Sec.\,\ref{sec:C} describes how our regularization strategy tackles the ``lost" kernel point phenomenon.
    \item Sec.\,\ref{sec:D} enumerates more segmentation results with class scores.
    \item \textbf{KPConv Method} video\footnote{\textit{\urlstyle{sf}\url{https://www.youtube.com/watch?v=uwuvp9mc_0o&t=19s}}} illustrates KPConv principle with animated diagrams, and shows some learned kernel deformations.
    \item \textbf{KPConv Results} video\footnote{\textit{\urlstyle{sf}\url{https://www.youtube.com/watch?v=_cFQxJorSAI}}}  shows indoor and outdoor scenes segmented by KP-FCNN.
\end{itemize}
\end{abstract}

\section{Network Architectures and Parameters}
\label{sec:A}

\begin{figure}[b]
    \centering
    \includegraphics[width=0.98\columnwidth, keepaspectratio=true]{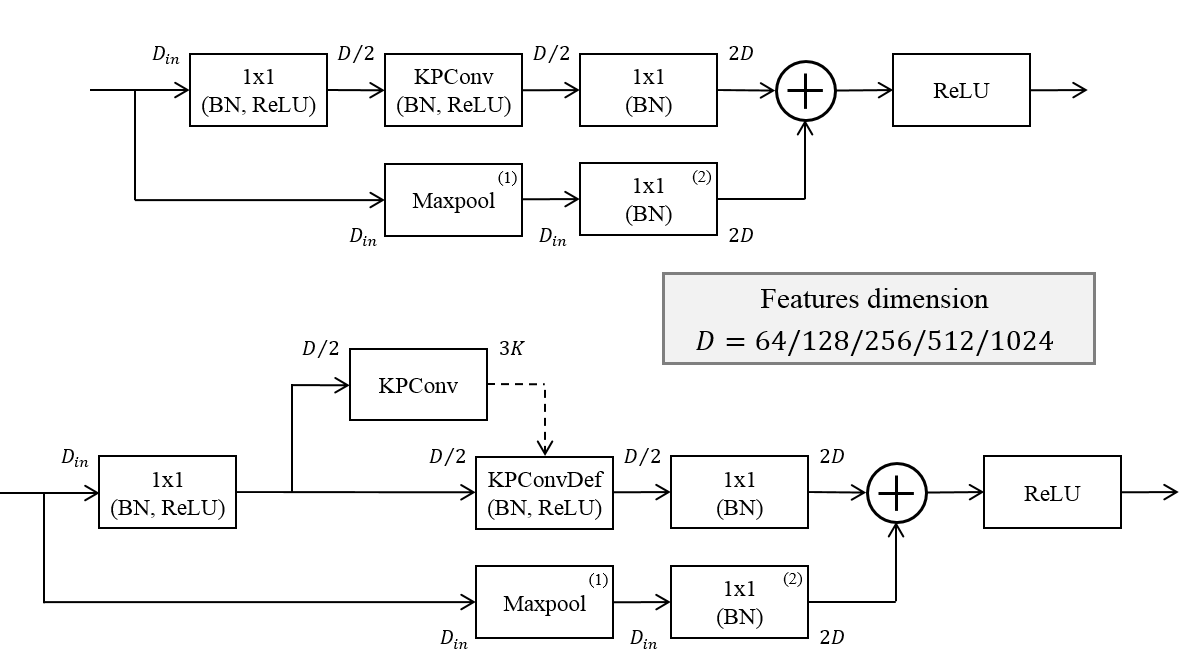}
    \caption{Convolutional blocks used in our architectures. Both rigid (top) and deformable (bottom) KPConv use resnet connections, batch normalization and leaky ReLU. Optional blocks: shortcut max pooling$^{(1)}$ is only needed for strided KPConv, and shortcut 1x1 convolution$^{(2)}$ is only needed when $D_{in} \neq 2D$.}
    \label{fig_blocks}
\end{figure}

\begin{figure*}[t]
    \vspace{1ex}
    \centering
    \includegraphics[width=0.98\textwidth, keepaspectratio=true]{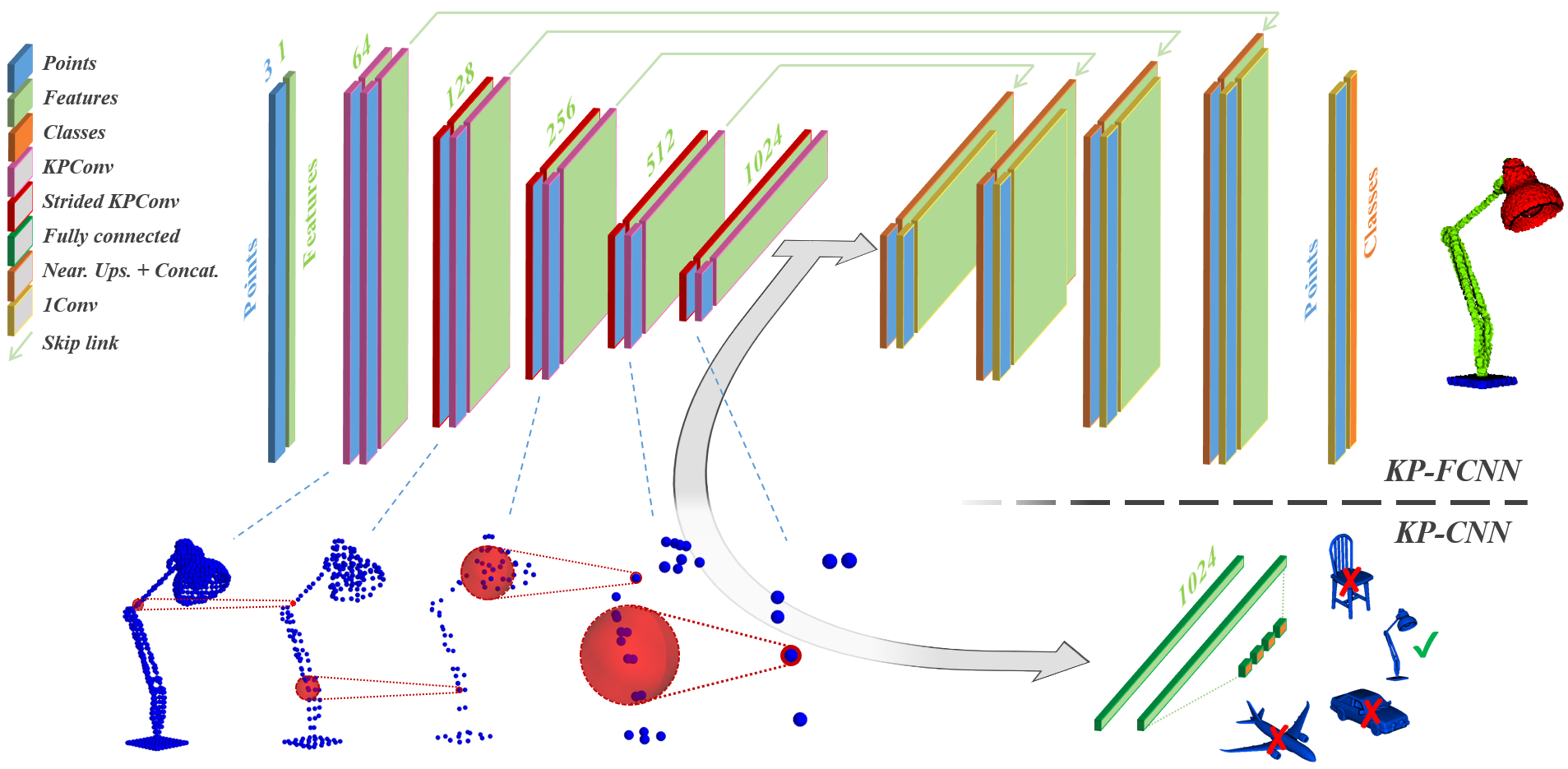}
    \caption{Illustration of our 2 network architectures for segmentation (top) and classification (bottom) of 3D point clouds. During a forward pass, features are transformed by consecutive operations (represented by edge colors) while points are fed to each layer as a support structure guiding the operations.}
    \label{fig_networks}
    \vspace{1ex}
\end{figure*}

As explained in the main paper, our architectures are built with convolutional blocks, designed like bottleneck ResNet blocks \cite{he2016deep}. This is the case whether we use a normal or strided KPConv, with rigid or deformable kernels. Figure \ref{fig_blocks} describes these blocks, and Figure \ref{fig_networks} our two network architectures made from them. In Figure \ref{fig_networks}, we show an example of point cloud from ModelNet40 dataset, subsampled at every layer. It illustrates how the convolution radius (red sphere) grows proportionally to the subsampling grid size. In all our experiments with deformable KPConv, we use deformable kernels in the last 5 convolutional blocks (2$^{nd}$ block from layer 3, and both block from layer 4 and 5). The green number above layers in Figure \ref{fig_networks} are the feature dimensions used in our blocks ($D$ in Figure \ref{fig_blocks}).

Our layers process point clouds of variable sizes, so we cannot stack them along a new ``batch'' dimension. We thus stack our point and feature tensors along their first dimension (number of points). As the neighbor and pooling indices do not point from one input cloud to another, each batch element is processed independently without any implementation trick. We only need to keep track of the batch element indices in order to define the global pooling of \textbf{KP-CNN}. Since the number of points can vary a lot, we use a variable batch size by selecting as many elements as possible until a certain number of total of batch points is reached. This limit is chosen so that the average batch size correspond to the target batch size. A very similar batch strategy has already been described by \cite{hermosilla2018monte}.

\noindent
\textbf{KP-CNN training}. We use a Momentum gradient Descent optimizer to minimize a cross-entropy loss, with a batch size of 16, a momentum of $0.98$ an initial learning rate of $10^{-3}$. Our learning rate is scheduled to decrease exponentially, and we choose the exponential decay to ensure it is divided by 10 every 100 epochs. A $0.5$ probability dropout is used in the final fully connected layers. The network converges in 200 epochs. In the case of deformable kernels, the regularization loss is added to the output loss with a multiplicative factor of $0.1$.

\noindent
\textbf{KP-FCNN training}. We also use a Momentum gradient Descent optimizer to minimize a point-wise cross-entropy loss, with a batch size of 10, a momentum of $0.98$ an initial learning rate of $10^{-2}$. The same learning rate schedule is used and no dropout is used. Among all experiments, the network needs 400 epochs at most to converge. For real scene segmentation, we can generate any number of input spheres, so we define an epoch as 500 optimizer steps, which is equivalent to 5000 spheres seen by the network. The same deformable regularization loss is used.

\begin{table}[b]
\setlength\tabcolsep{0.5pt}
\begin{center}
\begin{tabular}{ L{1.5cm} C{1.0cm} C{1.3cm} C{1.3cm} C{1.3cm}  C{1.3cm} }
 & & MN40 & SNP & Scannet & Sem3D \TBstrut\\
\hline
\multicolumn{2}{c}{Avg pts/elem} & $6800$	& $2370$	& $8950$	& $3800$	\Tstrut\\
\multicolumn{2}{c}{Avg pts/batch} & $109\mathrm{K}$	& $38\mathrm{K}$	& $90\mathrm{K}$	& $38\mathrm{K}$	\Bstrut\\
\hline
\multirow{2}{*}{Params} & \textit{rigid} & $14.3\mathrm{M}$	& $14.2\mathrm{M}$	& $14.1\mathrm{M}$	& $14.1\mathrm{M}$	\Tstrut\\
 & \textit{deform}  & $15.2\mathrm{M}$	& $15.0\mathrm{M}$	& $14.9\mathrm{M}$	& $14.9\mathrm{M}$ \Bstrut\\
\hline
Training & \textit{rigid}	& $3.5$	& $5.5$	& $4.3$	& $8.8$	\Tstrut\\
(batch/s) & \textit{deform}	& $3.1$	& $4.3$	& $3.9$	& $7.1$	\Bstrut\\
\hline
Inference  & \textit{rigid}	& $8.7$	& $16.7$	& $9.3$	& $17.5$	\Tstrut\\
(batch/s) & \textit{deform}	& $8.0$	& $12.2$	& $8.1$	& $15.0$	\Bstrut\\
\hline
\end{tabular}
\end{center}
\caption{Model statistics on 4 datasets: ModelNet40, ShapeNetPart, Scannet, Semantic3D.}
\label{Table_speeds}
\end{table}

\noindent
\textbf{Model sizes and speeds}. Table \ref{Table_speeds} shows the statistics of our models on different datasets. First we notice that KP-FCNN and KP-CNN have similar number of parameters, because the decoder part of KP-FCNN only involves light 1x1 convolution. We see that the running speeds are different from one dataset to another, which is not surprising. Indeed, the number of operations performed during a forward pass of our network depends on the number of points of the current batch, and the maximum number of neighbors of these points. Our models have been prototyped with a RTX 2080Ti in this experiment, which explains the slight difference with the Titan Xp used in the main paper.

\section{Kernel Points Initialization}
\label{sec:B}

Our KPConv operates in a ball, and requires kernel points regularly placed in this domain. There is no obvious regular disposition of points in a sphere, so we chose to solve this issue by translating it into an optimization problem. The problem is simple, we want the $K$ points $\widetilde{x}_k$ to be as far from each other as possible inside a given sphere. We thus assign a repulsive potential to each point:

\begin{equation} \label{eq:9}
    \forall x \in \mathbb{R}^3,\quad E_k^{rep}(x)= \frac{1}{\left\Vert x-\widetilde{x}_k \right\Vert}
\end{equation}

\noindent
And add an attractive potential to the sphere center to avoid them diverging indefinitely:

\begin{equation} \label{eq:10}
    \forall x \in \mathbb{R}^3,\quad E^{att}(x)= \left\Vert x \right\Vert^2
\end{equation}

\noindent
The problem then consists of minimizing the global energy:

\begin{equation} \label{eq:11}
    E^{tot} = \sum_{k<K} \left( E^{att}(\widetilde{x}_k) + \sum_{l\neq k} E_k^{rep}(\widetilde{x}_l) \right)
\end{equation}

\begin{figure}[t]
    \vspace{1ex}
    \centering
    \includegraphics[width=0.95\columnwidth, keepaspectratio=true]{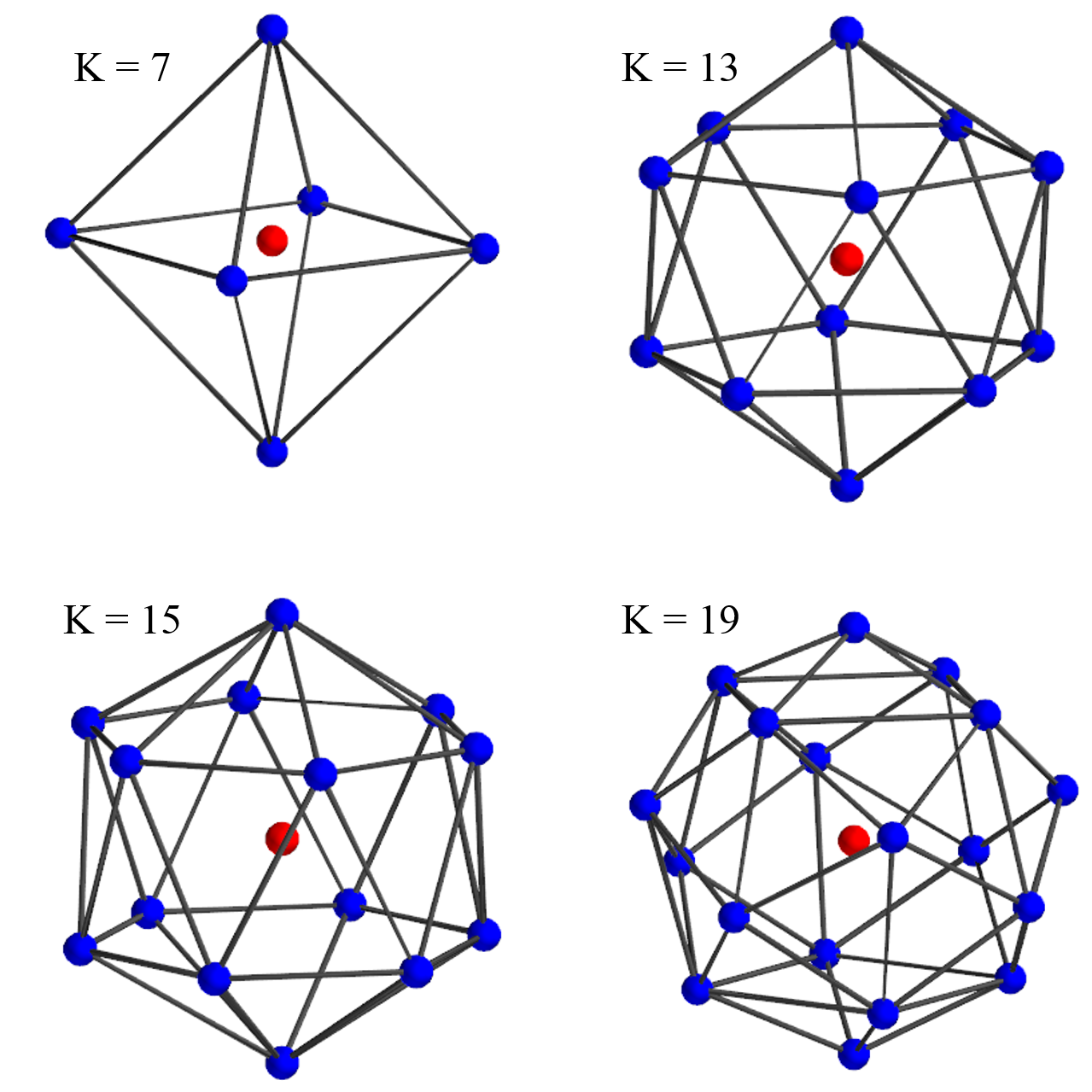}
    \caption{Illustration of the kernel points in stable dispositions.}
    \label{fig_kernels}
    \vspace{1ex}
\end{figure}

\begin{table}[b]
\begin{center}
\begin{tabular}{ C{0.5cm} C{2.0cm} C{3.0cm}}
\multirow{2}{*}{$K$} & disposition & groups along \\
 & name & symmetrical axis  \Bstrut\\
\hline
$5$  & Tetrahedron &  \Tstrut\\
$7$  & Octahedron & 1-4-1 \\
$13$ & Icosahedron & 1-5-5-1 \\
$15$ & - & 1-6-6-1 \\
$18$ & - & 1-5-5-5-1 \\
$19$ & - & 1-4-4-4-4-1 \\
$21$ & - & 1-6-6-6-1 \\
$25$ & - & 4-4-4-4-4-4  \Bstrut\\
\hline
\end{tabular}
\end{center}
\caption{Stable dispositions of the kernel point positions when the center point is fixed. If a disposition has an axis of symmetry, we describe it by the successive groups of points sharing a plane perpendicular to this axis.}
\label{Table_disposition}
\end{table}

The solution is found by gradient descent with the points initialized randomly and some optional constraints. In our case, we fix one of the points at the center of the sphere. For some values of $K$ (listed in Table \ref{Table_disposition}), the points converge to a unique stable disposition. Those stable dispositions are in fact regular polyhedrons. Each polyhedron can be described by grouping points sharing a plane perpendicular to the polyhedron symmetrical axis. For a better understanding, some of these dispositions are shown in Figure \ref{fig_kernels}.

In every layer of KP-CNN and KP-FCNN, the points locations are rescaled from the chosen stable disposition to the appropriate radius and randomly rotated. Note that $E^{tot}$ can also be used as a regularization loss in KP-CNN, when the kernel point positions are trained.

\section{Effect of the Kernel Point Regularization}
\label{sec:C}

When we designed deformable KPConv, we first used a straightforward adaptation of image deformable convolutions, but the network had very poor performances. We investigated the kernel deformations after the network convergence and noticed that the kernel points were often pulled away from the input points. This phenomenon comes from the sparse nature of point clouds, there is empty space around the points. We remind that the shifts are predicted by the network, thus, they depend on the input shape. 

For a particular input during training, if a kernel point is shifted away from the input points, then the gradient of its shift $\Delta_k(x)$ is null. It is thus ``lost" by the network and remains away for similar input shapes. Because of the stochastic nature of the network optimizer, this happens for many input shapes during convergence. 

\begin{figure}[b]
    \centering
    \includegraphics[width=0.95\columnwidth, keepaspectratio=true]{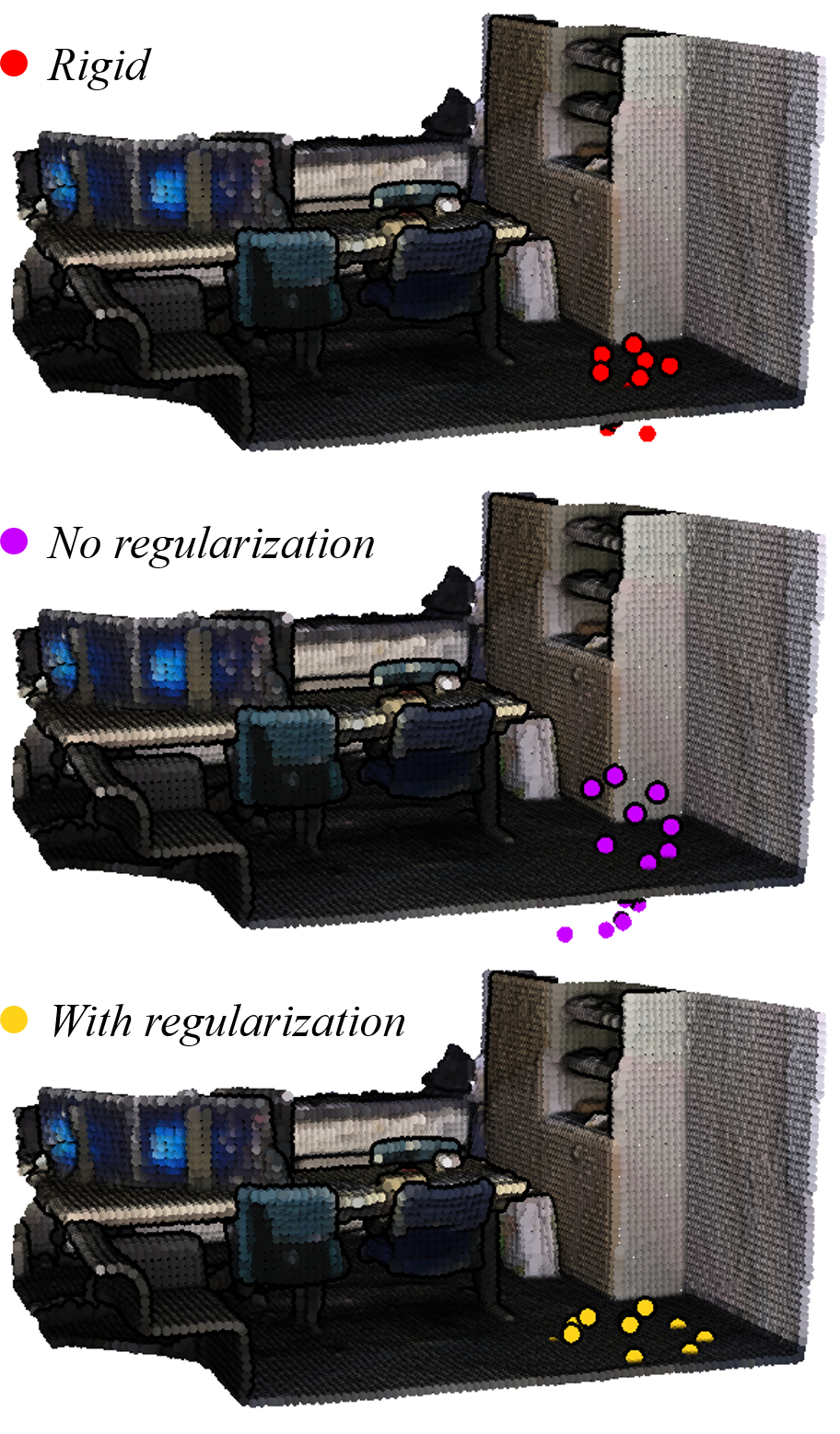}
    \vspace{1ex}
    \caption{Illustration of the deformations learned by a KPConv network with or without regularization.}
    \label{fig_regu}
\end{figure}

Figure \ref{fig_regu} illustrates ``lost" kernel points on the example of a room floor. First we see the rigid kernel in red, its scale gives an idea of the kernel points influence range. In the middle, the purple points depict a deformed kernel predicted by a network without any regularization loss. Most purple points are far from the floor plane and thus ``lost". 

Our regularization strategy, described in the main paper, prevents this phenomenon, as shown in the bottom of Figure \ref{fig_regu}. We can notice that our regularization strategy does not only prevent the ``lost" kernel points. It also helps to maximize the number of active kernel points in KPConv (those with input points in range). Almost every yellow point is close to the floor plane.

\section{More Segmentation Results}
\label{sec:D}

In this section, we provide more details on our segmentation experiments, for benchmarking purpose with future works. We give class scores for our experiments on ShapeNetPart (Table \ref{Table_ShapeNetPart}) and S3DIS (Tables \ref{Table_S3DIS_area5} and \ref{Table_S3DIS_kfold}) dataset. Scannet \cite{dai2017scannet}, Semantic3D \cite{hackel2017semantic3d} and NPM3D \cite{roynard2018paris} are online benchmarks, the class scores can be found on their respective website.

\begin{table*}[b]
\setlength\tabcolsep{0.5pt}
\begin{small}
\begin{center}
\begin{tabular}{L{2.5cm} | C{0.9cm} | C{0.9cm} | *{16}{C{0.75cm}}}
\multirow{2}{*}{Method} & class & inst. & \multirow{2}{*}{aero} & \multirow{2}{*}{bag} & \multirow{2}{*}{cap} & \multirow{2}{*}{car} & \multirow{2}{*}{chair} & \multirow{2}{*}{ear} & \multirow{2}{*}{guit} & \multirow{2}{*}{knif} & \multirow{2}{*}{lamp} & \multirow{2}{*}{lapt} & \multirow{2}{*}{moto} & \multirow{2}{*}{mug} & \multirow{2}{*}{pist} & \multirow{2}{*}{rock} & \multirow{2}{*}{skate} & \multirow{2}{*}{table} \\
 & avg. & avg. & & & & & & & & & & & & & & & & \Bstrut\\
\hline
Kd-Net \cite{klokov2017escape} 	& $77.4$	& $82.3$	& $80.1$	& $74.6$	& $74.3$	& $70.3$	& $88.6$	& $73.5$	& $90.2$	& $87.2$	& $81.0$	& $94.9$	& $57.4$	& $86.7$	& $78.1$	& $51.8$	& $69.9$	& $80.3$	\Tstrut\\
SO-Net  \cite{li2018so}	& $81.0$	& $84.9$	& $82.8$	& $77.8$	& $88.0$	& $77.3$	& $90.6$	& $73.5$	& $90.7$	& $83.9$	& $82.8$	& $94.8$	& $69.1$	& $94.2$	& $80.9$	& $53.1$	& $72.9$	& $83.0$	\\
PCNN by Ext \cite{atzmon2018point}	& $81.8$	& $85.1$	& $82.4$	& $80.1$	& $85.5$	& $79.5$	& $90.8$	& $73.2$	& $91.3$	& $86.0$	& $85.0$	& $95.7$	& $73.2$	& $94.8$	& $83.3$	& $51.0$	& $75.0$	& $81.8$	\\
PointNet++ \cite{qi2017pointnet++}	& $81.9$	& $85.1$	& $82.4$	& $79.0$	& $87.7$	& $77.3$	& $90.8$	& $71.8$	& $91.0$	& $85.9$	& $83.7$	& $95.3$	& $71.6$	& $94.1$	& $81.3$	& $58.7$	& $76.4$	& $82.6$	\\
SynSpecCNN \cite{yi2017syncspeccnn}	& $82.0$	& $84.7$	& $81.6$	& $81.7$	& $81.9$	& $75.2$	& $90.2$	& $74.9$	& $\mathbf{93.0}$	& $86.1$	& $84.7$	& $95.6$	& $66.7$	& $92.7$	& $81.6$	& $60.6$	& $\mathbf{82.9}$	& $82.1$	\\
DGCNN \cite{wang2018dynamic}	& $82.3$	& $85.1$	& $84.2$	& $83.7$	& $84.4$	& $77.1$	& $90.9$	& $78.5$	& $91.5$	& $87.3$	& $82.9$	& $96.0$	& $67.8$	& $93.3$	& $82.6$	& $59.7$	& $75.5$	& $82.0$	\\
SpiderCNN \cite{xu2018spidercnn}	& $82.4$	& $85.3$	& $83.5$	& $81.0$	& $87.2$	& $77.5$	& $90.7$	& $76.8$	& $91.1$	& $87.3$	& $83.3$	& $95.8$	& $70.2$	& $93.5$	& $82.7$	& $59.7$	& $75.8$	& $82.8$	\\
SubSparseCNN \cite{graham20183d}	& $83.3$	& $86.0$	& $84.1$	& $83.0$	& $84.0$	& $80.8$	& $\mathbf{91.4}$	& $78.2$	& $91.6$	& $\mathbf{89.1}$	& $85.0$	& $95.8$	& $73.7$	& $95.2$	& $84.0$	& $58.5$	& $76.0$	& $82.7$	\\
SPLATNet \cite{su2018splatnet}	& $83.7$	& $85.4$	& $83.2$	& $84.3$	& $89.1$	& $80.3$	& $90.7$	& $75.5$	& $92.1$	& $87.1$	& $83.9$	& $96.3$	& $75.6$	& $\mathbf{95.8}$	& $83.8$	& $64.0$	& $75.5$	& $81.8$	\\
PointCNN \cite{li2018pointcnn}	& $84.6$	& $86.1$	& $84.1$	& $86.5$	& $86.0$	& $80.8$	& $90.6$	& $79.7$	& $92.3$	& $88.4$	& $\mathbf{85.3}$	& $96.1$	& $77.2$	& $95.3$	& $84.2$	& $64.2$	& $80.0$	& $83.0$	\\
FlexConv \cite{groh2018flex}	& $85.0$	& $84.7$	& $83.6$	& $\mathbf{91.2}$	& $\mathbf{96.7}$	& $79.5$	& $84.7$	& $71.7$	& $92.0$	& $86.5$	& $83.2$	& $\mathbf{96.6}$	& $71.7$	& $95.7$	& $86.1$	& $\mathbf{74.8}$	& $81.4$	& $\mathbf{84.5}$	\Bstrut\\
\hline
KPConv \textit{rigid}	& $85.0$	& $86.2$	& $83.8$	& $86.1$	& $88.2$	& $\mathbf{81.6}$	& $91.0$	& $\mathbf{80.1}$	& $92.1$	& $87.8$	& $82.2$	& $96.2$	& $77.9$	& $95.7$	& $\mathbf{86.8}$	& $65.3$	& $81.7$	& $83.6$	\Tstrut\\
KPConv \textit{deform}	& $\mathbf{85.1}$	& $\mathbf{86.4}$	& $\mathbf{84.6}$	& $86.3$	& $87.2$	& $81.1$	& $91.1$	& $77.8$	& $92.6$	& $88.4$	& $82.7$	& $96.2$	& $\mathbf{78.1}$	& $\mathbf{95.8}$	& $85.4$	& $69.0$	& $82.0$	& $83.6$	\Bstrut\\
\hline
\end{tabular}
\end{center}
\end{small}
\caption{Segmentation mIoUs on ShapeNetPart.}
\label{Table_ShapeNetPart} 
\end{table*}

\begin{table*}[b]
\setlength\tabcolsep{0.5pt}
\begin{small}
\begin{center}
\begin{tabular}{L{2.4cm} | C{1.0cm} | C{1.0cm} | *{13}{C{0.9cm}}}
Method	 & mIoU & mRec & ceil.	 & floor	 & wall	 & beam	 & col.	 & wind.	 & door	 & chair	 & table	 & book.	 & sofa	 & board & clut.	\Bstrut\\
\hline
Pointnet \cite{qi2017pointnet}	& $41.1$	& $49.0$	& $88.8$	& $97.3$	& $69.8$	& $0.1$	& $3.9$	& $46.3$	& $10.8$	& $52.6$	& $58.9$	& $40.3$	& $5.9$	& $26.4$	& $33.2$	\Tstrut\\
SegCloud \cite{tchapmi2017segcloud}	& $48.9$	& $57.4$	& $90.1$	& $96.1$	& $69.9$	& $0.0$	& $18.4$	& $38.4$	& $23.1$	& $75.9$	& $70.4$	& $58.4$	& $40.9$	& $13.0$	& $41.6$	\\
Eff 3D Conv \cite{zhang2018efficient}	& $51.8$	& $68.3$	& $79.8$	& $93.9$	& $69.0$	& $0.2$	& $28.3$	& $38.5$	& $48.3$	& $71.1$	& $73.6$	& $48.7$	& $59.2$	& $29.3$	& $33.1$	\\
TangentConv \cite{tatarchenko2018tangent}	& $52.6$	& $62.2$	& $90.5$	& $97.7$	& $74.0$	& $0.0$	& $20.7$	& $39.0$	& $31.3$	& $69.4$	& $77.5$	& $38.5$	& $57.3$	& $48.8$	& $39.8$	\\
RNN Fusion \cite{ye20183d}	& $57.3$	& $63.9$	& $92.3$	& $\mathbf{98.2}$	& $79.4$	& $0.0$	& $17.6$	& $22.8$	& $62.1$	& $74.4$	& $80.6$	& $31.7$	& $66.7$	& $62.1$	& $56.7$	\\
SPGraph \cite{landrieu2018large}	& $58.0$	& $66.5$	& $89.4$	& $96.9$	& $78.1$	& $0.0$	& $\mathbf{42.8}$	& $48.9$	& $61.6$	& $84.7$	& $75.4$	& $69.8$	& $52.6$	& $2.1$	& $52.2$	\\
ParamConv \cite{wang2018deep}	& $58.3$	& $67.1$	& $92.3$	& $96.2$	& $75.9$	& $\mathbf{0.3}$	& $6.0$	& $\mathbf{69.5}$	& $63.5$	& $66.9$	& $65.6$	& $47.3$	& $68.9$	& $59.1$	& $46.2$	\Bstrut\\
\hline
KPConv \textit{rigid}	& $65.4$	& $70.9$	& $92.6$	& $97.3$	& $81.4$	& $0.0$	& $16.5$	& $54.5$	& $\mathbf{69.5}$	& $90.1$	& $80.2$	& $74.6$	& $66.4$	& $63.7$	& $58.1$	\Tstrut\\
KPConv \textit{deform}	& $\mathbf{67.1}$	& $\mathbf{72.8}$	& $\mathbf{92.8}$	& $97.3$	& $\mathbf{82.4}$	& $0.0$	& $23.9$	& $58.0$	& $69.0$	& $\mathbf{91.0}$	& $\mathbf{81.5}$	& $\mathbf{75.3}$	& $\mathbf{75.4}$	& $\mathbf{66.7}$	& $\mathbf{58.9}$	\Bstrut\\
\hline
\end{tabular}
\end{center}
\end{small}
\caption{Semantic segmentation IoU scores on S3DIS \textit{Area-5}. Additionally, we give the mean class recall, a measure that some previous works call mean class accuracy.}
\label{Table_S3DIS_area5} 
\end{table*}

\begin{table*}[t]
\setlength\tabcolsep{0.5pt}
\begin{small}
\begin{center}
\begin{tabular}{L{2.4cm} | C{1.0cm} | C{1.0cm} | *{13}{C{0.9cm}}}
Method	 & mIoU & mRec & ceil.	 & floor	 & wall	 & beam	 & col.	 & wind.	 & door	 & chair	 & table	 & book.	 & sofa	 & board & clut.	\Bstrut\\
\hline
Pointnet \cite{qi2017pointnet}	& $47.6$	& $66.2$	& $88.0$	& $88.7$	& $69.3$	& $42.4$	& $23.1$	& $47.5$	& $51.6$	& $42.0$	& $54.1$	& $38.2$	& $9.6$	& $29.4$	& $35.2$	\Tstrut\\
RSNet \cite{huang2018recurrent}	& $56.5$	& $66.5$	& $92.5$	& $92.8$	& $78.6$	& $32.8$	& $34.4$	& $51.6$	& $68.1$	& $60.1$	& $59.7$	& $50.2$	& $16.4$	& $44.9$	& $52.0$	\\
SPGraph \cite{landrieu2018large}	& $62.1$	& $73.0$	& $89.9$	& $95.1$	& $76.4$	& $62.8$	& $47.1$	& $55.3$	& $68.4$	& $\mathbf{73.5}$	& $\mathbf{69.2}$	& $63.2$	& $45.9$	& $8.7$	& $52.9$	\\
PointCNN \cite{li2018pointcnn}	& $65.4$	& $75.6$	& $\mathbf{94.8}$	& $\mathbf{97.3}$	& $75.8$	& $63.3$	& $51.7$	& $58.4$	& $57.2$	& $71.6$	& $69.1$	& $39.1$	& $61.2$	& $52.2$	& $58.6$	\Bstrut\\
\hline
KPConv \textit{rigid}	& $69.6$	& $78.1$	& $93.7$	& $92.0$	& $82.5$	& $62.5$	& $49.5$	& $65.7$	& $\mathbf{77.3}$	& $57.8$	& $64.0$	& $68.8$	& $71.7$	& $60.1$	& $59.6$	\Tstrut\\
KPConv \textit{deform}	& $\mathbf{70.6}$	& $\mathbf{79.1}$	& $93.6$	& $92.4$	& $\mathbf{83.1}$	& $\mathbf{63.9}$	& $\mathbf{54.3}$	& $\mathbf{66.1}$	& $76.6$	& $57.8$	& $64.0$	& $\mathbf{69.3}$	& $\mathbf{74.9}$	& $\mathbf{61.3}$	& $\mathbf{60.3}$	\Bstrut\\
\hline
\end{tabular}
\end{center}
\end{small}
\caption{Semantic segmentation IoU scores on S3DIS \textit{k-fold}. Additionally, we give the mean class recall, a measure that some previous works call mean class accuracy.}
\label{Table_S3DIS_kfold} 
\end{table*}

\end{document}